\documentclass[sigconf]{acmart}

\usepackage{booktabs} 

\setcopyright{rightsretained}

\acmDOI{}

\acmISBN{}

\acmConference[WWW'21 KMECommerce Workshop]{The Web Conference - Knowledge Management in e-Commerce Workshop}{April 2021}{Ljubljana, Slovenia}
\acmBooktitle{Knowledge Management in e-Commerce Workshop at the Web Conference 2021 (WWW '21)}
\acmYear{2021}
\copyrightyear{2021}

\acmArticle{}
\acmPrice{15.00}

\begin{document}
\title{Graph Neural Networks for Inconsistent Cluster Detection in Incremental Entity Resolution}

\author{Robert A. Barton}
\orcid{1234-5678-9012}
\affiliation{%
  \institution{Amazon}
  \city{New York}
  \state{NY}
  \country{USA}
}
\email{rab@amazon.com}

\author{Tal Neiman}
\affiliation{%
  \institution{Amazon}
  \city{New York}
  \state{NY}
 \country{USA}
}
\email{talneim@amazon.com}

\author{Changhe Yuan}
\affiliation{%
  \institution{Amazon}
  \city{New York}
  \state{NY}
  \country{USA}}
\email{ychanghe@amazon.com}

\renewcommand{\shortauthors}{R. Barton et al.}

\begin{abstract}
Online stores often utilize product relationships such as bundles and substitutes to improve their catalog quality and guide customers through myriad choices. Entity resolution using pairwise product matching models offers a means of inferring relationships between products. In mature data repositories, the relationships may be mostly correct but require incremental improvements owing to errors in the original data or in the entity resolution system.  It is critical to devise incremental entity resolution (IER) approaches for improving the health of relationships.  However, most existing research on IER focuses on the addition of new products or information into existing relationships. Little research has been done for detecting low quality within current relationships and ensuring that incremental improvements affect only the unhealthy relationships.  


This paper fills the void in IER research by developing a novel method for identifying inconsistent clusters (IC), existing groups of related products that do not belong together. We propose to treat the identification of inconsistent clusters as a supervised learning task which predicts whether a graph of products with similarities as weighted edges should be partitioned into multiple clusters. In this case, the problem becomes a classification task on weighted graphs and represents an interesting application area for modern tools such as Graph Neural Networks (GNNs). We demonstrate that existing Message Passing neural networks perform well at this task, exceeding traditional graph processing techniques. We also develop a novel message aggregation scheme for Message Passing Neural Networks that further improves the performance of GNNs on this task. We apply the model to synthetic datasets, a public benchmark dataset, and an internal application. Our results demonstrate the value of graph classification in IER and the ability of graph neural networks to develop useful representations for graph partitioning.  

\end{abstract}

%
%



\maketitle



\section{Introduction}
Entity resolution, the problem of grouping related entities together, is an important problem in computer science with many application domains, such as databases (when linking records from different databases that represent the same object) or natural language processing (disambiguating nouns that refer to the same entity)\cite{fellegi1969theory,cohen2002learning,bhattacharya2007collective}.  Much of the research in entity resolution focuses on static systems where the resolution is performed as a one-time exercise.  In many applications, however, a large repository of existing relationships must be continually updated in order to maintain correct groupings of entities.  Incremental entity resolution, or dynamic entity resolution, studies the problem of how to update repositories of existing relationships over time \cite{whang2010entity, gruenheid2014incremental, nentwig2018incremental}.  Studies of incremental entity resolution typically focus on how to adapt to new entities and information as they are received.  However, another important goal in practice is to make the resolved entities incrementally better while not decreasing quality \cite{wang2016clustering}.  Where cluster quality is important, we must detect incorrect clusters of relationships with high precision to ensure that healthy clusters are maintained.  Incorrect clusters of related entities can arise, for example, from poor quality in the original data, prior incorrect clustering decisions, changing item data, or malicious actors that fool existing entity resolution systems.    We refer to clusters or families of incorrectly resolved entities as inconsistent clusters (IC).


As an example of where Inconsistent Cluster detection is important, consider entity resolution in the domain of e-commerce products.  Stores such as Amazon, AliExpress, WalMart, and eBay provide huge selections of products for customers. These platforms often utilize product relationships to link relevant products together or link them to customers.  Examples of product relationships include duplicates, complements, substitutes, and accessories.  Many entity resolution approaches have been proposed for these product matching problems ~\cite{konda2016magellan,mudgal2018deep,ristoski2018machine}.  Another useful but less researched relationship is product variations, which refer to families of products that are functionally the same but differ in specific attributes. As an example, shoes that are identical in all aspects other than size and/or color are variations of one another and are shown to customers on a single page to simplify selection. In this case, the entire resolved cluster is shown to the customer, and the inclusion of any shoe that does not belong will make the whole family inconsistent.  In this case, IC defects are directly visible to customers and are critical to fix.  More generally, inconsistent clusters of items can be detrimental in any entity resolution system.  

Conventional entity resolution methods were not developed to detect ICs with high precision for IER.  In the conventional approach, a matching step generates the probability for a relationship between all pairs of items. Matching is followed by a clustering step, wherein clusters are formed by applying a graph processing algorithm such as transitive closure to all edges above a threshold value \cite{bhattacharya2007collective, hassanzadeh2009framework}.  To determine whether existing clusters should be divided, one could apply matching and clustering within all existing families.  However, it is possible for pairwise scores to conflict \cite{reas2018superpart}, leading to incorrect cluster formation and preventing existing Inconsistent Families from being detected.  It has been observed that different clustering algorithms have different strengths and weaknesses \cite{hassanzadeh2009framework}.  We propose instead a classification task to evaluate cluster quality that generates a score for any given cluster being inconsistent.  In addition to allowing us to escape the weaknesses of any particular graph processing algorithm, this approach has the benefit that the score can be tuned to ensure high precision identification of ICs.   Before either forming a new cluster or disaggregating an existing cluster, IC detection can be used to ensure high confidence that every cluster is incrementally improved.  This ability is especially useful when the quality of each individual cluster is important such as in the product variations problem above.

We formulate the IC detection problem as a graph classification problem on weighted undirected graphs, where edges represent the similarities between each node.  We then develop a binary graph classifier to predict whether the candidate items jointly form a consistent cluster.  Although we have motivated it from an entity resolution perspective, the problem represents a rich area of study for graph classification since the data are well represented in graph form with node features associated with entities and pairwise similarity scores associated with edges.  We apply graph neural networks \cite{gori2005new, kipf2016semi, velivckovic2017graph, hamilton2017inductive, chen2018supervised} to the problem of scoring a group of entities for whether they belong together.  We demonstrate that existing Graph Neural Network architectures perform well at graph classification on weighted graphs, and we develop a novel message representation that concatenates multiple node aggregation functions together to improve performance.  We tested our methods in a variety of problems, including an e-commerce application,  open data, and synthetic datasets, and demonstrate the general superiority of the proposed Graph Neural Network approach.

%
                                                                                                                                                                                  
To summarize, we make the following major contributions in this work. First, we introduce the Inconsistent Cluster detection problem as an important problem in Incremental Entity Resolution and demonstrate its usefulness on a real-world product variations dataset. Second, we develop a Graph Neural Network-based binary classifier for solving the Inconsistent Cluster detection problem. We demonstrate a novel message aggregation scheme that significantly improves the performance of a GNN classifier on this task, and show that its performance exceeds state of the art.  Third, our extensive evaluations shed light on the superior performance of the proposed method against several existing methods. The results show that existing methods may still work well when positive and negative edges are reasonably separated. But when the scores become more mixed and noisy, the advantage of our graph classifier is clear.  Finally, we demonstrate that Graph Neural Networks can learn robust representations of weighted graphs for reasoning about partitioning decisions, which we believe will be a valuable area of study for graph representation learning.

\section{Background}

In this section, we review several areas of research that are closely related to this work.

\subsection{Product matching}

Much existing research on product relationships in e-commerce focuses on product matching, e.g., duplicates, complements and substitutes. Product matching is usually formulated as an entity resolution problem ~\cite{fellegi1969theory,cohen2002learning}.  Konda et al. described an end-to-end entity resolution system that includes steps from data preprocessing to product matching~\cite{konda2016magellan}. More recently, many deep learning-based methods have been developed for entity resolution. Mudgal et al. evaluated several deep learning-based approaches for entity matching and found that deep learning-based methods do not outperform existing approaches in more structured domains, but can significantly outperform them on textual and noisy data~\cite{mudgal2018deep}. Shah et al. discussed two approaches for product matching, including a classification-based shallow neural network and a similarity-based on deep siamese network~\cite{shah2018neural}. Ristoski et al. used both word embeddings from neural language models and image embeddings from a deep convolutional neural network for product matching~\cite{ristoski2018machine}.  Li et al. studied using product titles and attributes for product matching and evaluate their method in two application scenarios, either as a classification problem or as a ranking problem~\cite{li2020deep}. Datasets that are specifically targeted for product matching have been published. For example, a WDC dataset has been published by \citet{primpeli2019wdc} for large-scale product matching. Its English language subset consists of 20 million pairs of duplicate products that were extracted from 43 thousand e-commerce websites. 

Another closely related area of research is product graphs~\cite{hammack2011handbook,dong2018challenges}. A product graph is a framework to capture all aspects of entities in a unified representation, including both product attributes and the relations between them. Such a unified view is beneficial because identifying relevant entity attributes and learning relationships are tightly coupled and can benefit from being treated jointly.

\subsection{Incremental entity resolution (IER)}
IER has been studied previously with different objectives.  \citet{whang2010entity, whang2014incremental} address the problem from the perspective of evolving matching rules, and point out that the IC problem arises in cases where a stricter matching rule requires all existing clusters to be reevaluated.  \citet{gruenheid2014incremental} discusses IER following record insertion, deletion, and change and allows existing clusters to split apart in response to these operations.  The main difference between these works and our own is that we focus on the precision of updates to individual clusters rather than relying on the ER clustering algorithm to cluster optimally.  These goals are important in an IER system that may already have relatively high cluster quality that needs to be preserved.  

Repairing existing relationships has also been studied from an ER perspective \cite{wang2016clustering} via the creation of a provenance index that tracks how clustering has changed over time.  In this work, we identify clusters that need repairs with high precision when the provenance may not be known.  


\subsection{Collective entity resolution}

Collective entity resolution, broadly defined, treats clusters of products jointly in entity resolution \cite{bhattacharya2007collective}. Many existing approaches to collective entity resolution can be described as a two-step process: matching and clustering. The matching step can be solved with the methods reviewed in the previous section and outputs scores of all edges in a cluster. The clustering step can also be solved using many methods. One popular method is {\em transitive closure} (TC) \cite{bhattacharya2007collective, hassanzadeh2009framework}.  In this method, edges below a critical threshold $t$ are removed and connected components are computed for remaining edges.  {\em K-Core} is another common method for the clustering step. In graph theory, k-cores of the graph are connected components that remain after all vertices with degrees less than $k$ are recursively removed.  These methods can be evaluated on our task by whether they split a cluster into more than one partition after their application.  

Finally, a more recent method called {\em SuperPart} is a cluster scoring model \cite{reas2018superpart}. SuperPart directly formulates classification of a weighted graph as having one or multiple partitions as a binary classification problem.  The SuperPart model featurizes a graph of pairwise edge scores by calculating graph-level features such as the mean external edge (mean of the edges removed by transitive closure at a fixed threshold), the average edge score in the cluster, and the number of connected components after filtering out edges below various thresholds.  A random forest is used to extract a cluster-level score from the table of features that represents a likelihood for the cluster to remain as a single partition.  To our knowledge, SuperPart is the only prior algorithm that assigns scores to evaluate cluster quality.  We will use it as a baseline algorithm in this work.  

\subsection{Graph neural networks}

Graph neural networks (GNNs) have been an increasingly active research area in recent years \cite{hamilton2017representation}. GNNs represent relationships between a set of entities as a graph, allowing the incorporation of both entity-level features as well as edge-level features. Many real-world problems can be represented as graphs, such as social networks \cite{kipf2016semi}, chemical and biological compounds  \cite{duvenaud2015convolutional, gilmer2017neural, velivckovic2017graph} and product graphs \cite{chen2018supervised}. One widely applicable formulation of graph neural networks, Message Passing Neural Networks  \cite{gilmer2017neural}, uses message propagation to iteratively update embedding representations of nodes, such that the embeddings capture the common patterns between node features directly communicated via edge features. Graph representation learning has been studied extensively, e.g., \cite{gori2005new, kipf2016semi, velivckovic2017graph, hamilton2017inductive, chen2018supervised}; however, GNN research has focused more on learning node embeddings than on graph classification.  Gilmer et al. performed graph-level regression to predict 13 chemical properties on a large graph dataset, whereas here we have a graph-level classification task related to the graph partitioning \cite{gilmer2017neural} .  Chen et al. perform graph partitioning at the node level given a fixed number of partitions \cite{chen2018supervised}.  This task is similar to the classification task described here in that it involves graph partitioning, but is dissimilar in that we have weighted rather than binary edges and are classifying entire graphs rather than nodes.  Graph Neural Networks have been explored for entity linkage in knowledge bases in Neural Collective Entity Linking (NCEL) \cite{cao2018neural}, wherein the best candidate edge for a given node is found using an existing clustering.  Overall, graph neural network models are particularly compelling for entity resolution because of their flexibility: they can learn from both node features and edge features and make predictions for nodes, edges, or entire graphs.  To our knowledge, graph neural networks have not yet been applied to evaluating clusters in entity resolution, for which many applications still use the traditional approach of calculating similarity scores followed by algorithmic clustering approaches.



\section{Method}

\begin{figure*}[!t]
\centering
\includegraphics[width=5in]{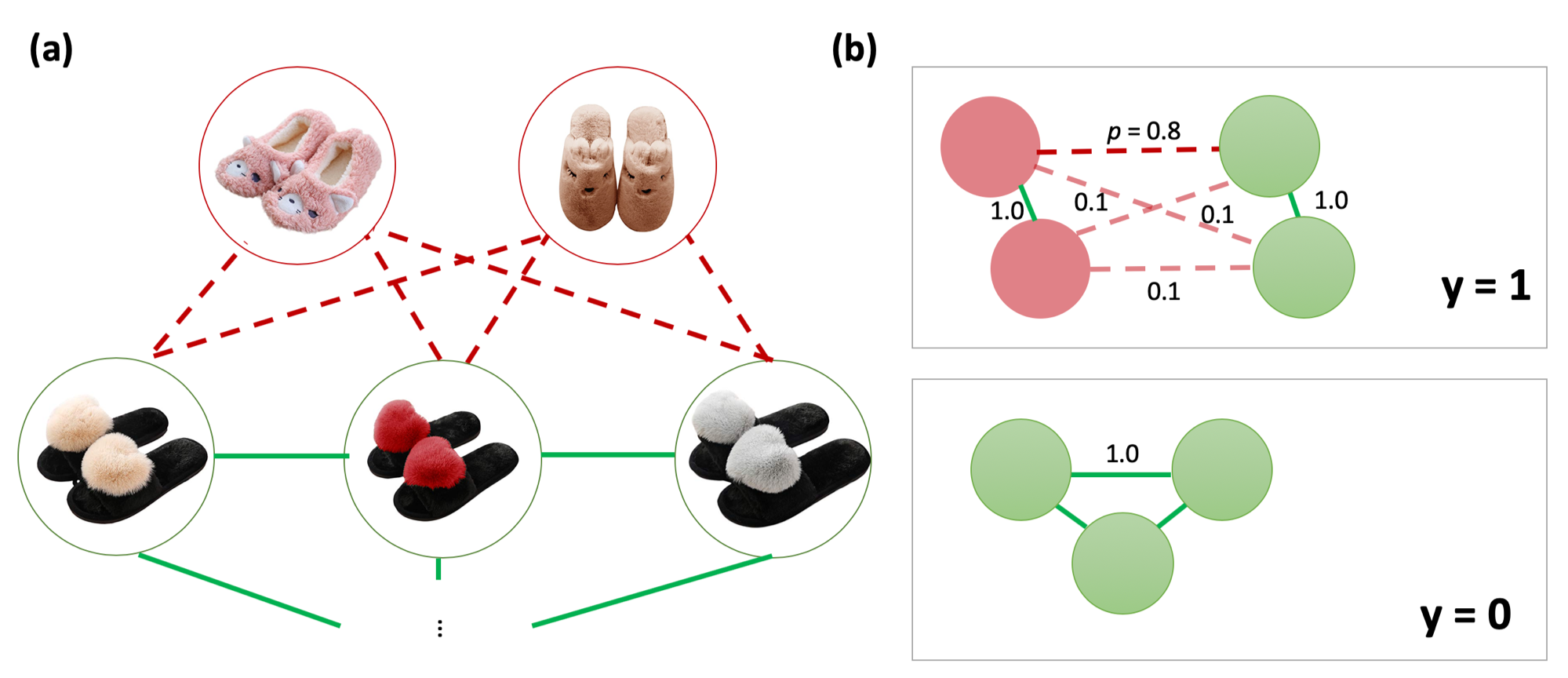}
\caption{(a) Example of an IC in the context of product variations.  Two items representing animal-themed slippers should be separated from traditional Womens' slippers.  (b) Generic binary graph classification problem.  Existing fully connected graphs of items contain both correct relationships (solid green lines), and incorrect relationships (dotted red lines) according to auditor labels.  The edge labels encode a partitioning of the graph into one subgraph in the case of healthy families ($y = 0$) or multiple subgraphs in the case of IC ($y = 1$).  The goal of the classifier is to find the correct graph label $y$ without knowing the true partitioning.  The classifier relies on a probability for edge existence $p$ derived from pairwise item similarity.  As in the IC family in (b), the task is challenging for algorithmic approaches because $p$ might vary across two true clusters.}
\label{fig:problemsetup}
\end{figure*}

\subsection{Problem}

Product variations are increasingly used by e-commerce websites to group similar products into families so that each family can be shown on a single detail page. Again taking shoes as an example, the shoes that share key aspects such as brand and style would be displayed on the same page; non-key aspects such as size and color are shows as selectable (dropdown) options. Such aggregated views of products allow customers to select specific products much more easily. Since all varied items are shown together, even a single item incorrectly grouped with others creates a visibly inconsistent family.  Importantly, stores such as WalMart, Amazon, and Shopify offer sellers the opportunity to upload information about variations directly, allowing for the possibility that incorrect clusters can arise as a result of input data rather than as the output of an entity resolution system. 

In this paper, we focus on the problem of detecting whether a family of items form a true cluster, which we define as the {\em Inconsistent Cluster} (IC) problem. Due to the need to treat product families collectively, IC detection can be much more challenging than simple pairwise entity matching. Imagine a family of products with 10 members, and only one of them does not belong. The family as a whole is clearly {\em defective}. However, if we consider the problem as pairwise product matching, a majority of the 45 product pairs (10 choose 2) are correct matches, with only 9 of them as mismatches. Although we can add a clustering step to further decide whether the family is defective or not, classic clustering approaches such as transitive closure or k-core may not work well for real world problems where noisy data are prevalent \cite{reas2018superpart}.   For example, one or two of the 9 scores for mismatched edges may be high.  

Importantly, because we use an independent algorithm to classify clusters into consistent and inconsistent clusters, our method can detect errors in the decision made by an entity resolution system.  This is an important contrast with other IER approaches because, as our data will show, clustering methods used in entity resolution do not always make correct decisions.  Even in this context of noisy pairwise scores, our method can be used to enforce high precision for resolved clusters.  

\subsection{Formulation}

We solve the IC detection problem by formulating it as a binary graph classification problem. Given an input graph $G(V,E)$, a signal  $y \in \{0, 1\}$ is to be reported for each graph (Figure \ref{fig:problemsetup}).  A graph with a negative label corresponds to a healthy family and a positive label corresponds to an IC.  As noted before, this graph classification task arises not only in finding existing families that are IC, but in confirming that items to be combined have a healthy score.  The nodes represent items and all candidate families of items are fully connected with edge scores $p_{ij}$ that represent item similarity between items $i$ and $j$.  

For maximum generality, in this work, we assume only that there is a means available to produce the pairwise node similarity feature $p$ between all candidate related items.  We create these pairwise edge scores for an internal e-commerce application, the inconsistent variation family detection problem, by using a random forest on edge features generated from each pair of records; these edge features are generated by a general record matching platform similar to \cite{de_bruin_j_2019_3559043}.  We also test the method on open data from \cite{reas2018superpart} which contains weighted graphs with ground truth partitionings.  Finally, we can test our method on synthetic data.  For this purpose, we adopt the Weighted Stochastic Block Model \cite{aicher2015learning}, wherein scores between two partitions come from one exponential family distribution, and scores within a given partition come from another.  

\subsection{Baseline Methods}
One common method for making partitioning decisions in entity resolution is transitive closure (TC).  In this method, edges below a critical threshold $t$ are removed and connected components are computed for remaining edges.  When enough edges fall below $t$ to divide a candidate family into two or more connected components, we predict $y = 1$ for the family-level label; otherwise, we predict $y = 0$.  Unlike other models studied here, this method does not provide a continuous score for how the cluster is likely to be classified.  However, we note that at a given threshold $t$, TC is a binary classifier which considers more families to be inconsistent as $t$ increases.  We can therefore create a precision-recall curve for the purpose of comparing to other graph classifiers by sweeping $t$.   To do this, we construct a family-level score $s_{i}^{TC} \equiv 1 - t_i^{min}$ where $t_i^{min}$ is minimum pairwise threshold at which family $i$ will have multiple connected components.  If $t_i^{min}$ is low ($s_{i}^{TC}$ is high), then the cluster breaks apart easily; however, if $t_i^{min}$ is high ($s_{i}^{TC}$ is low), even edges with high scores must be filtered out to divide the cluster.  We use the score $s_{i}^{TC}$ as a binary graph classification score for the purpose of constructing a precision / recall curve, where only families with score greater than or equal to a family-level threshold $t_f$ will be considered IC.

For k-core, we allow tunable $k \in [3,9] $, which is chosen in the validation job to optimize the $F_{1}$ metric we report.  The k-core algorithm recursively prunes nodes with degree less than $k$.  As for TC, we can calculate a PR-curve for k-core using the minimum pairwise threshold at which k-core would separate or reduce the size of a family.  

We also implement the SuperPart algorithm from \cite{reas2018superpart}, which explicitly solves the classification task of deciding whether a weighted graph should be broken into multiple partitions.  To our knowledge, SuperPart is the only existing graph classification method intended to classify clusters according to whether they should be partitioned.



\subsection{Proposed Model}

Going beyond prior work, we propose to solve the graph classification problem by applying graph neural network models (GNNs).  In particular, we adopt the message passing neural network (MPNN) framework outlined in \cite{gilmer2017neural} but propose a new message aggregation scheme that significantly improves the classification performance.  MPNNs can be described by their message passing, update, and readout functions.  Given a graph $G$ with node features $x_v$ and edge features $e_{vw}$, the MPNN works by aggregating ($\mathrm{AGG}$) messages $M_t$ at each node $v$ from neighboring nodes $N(v)$ at each timestep $t$.  These collected message $m_v^t$ can depend both on the neighboring node features as well as the edges.  The messages are then used to update the hidden state $h_v$ of the node according to an update function $U_t$.  Finally, after $T$ timesteps, the graph label is read from all node states according to a readout function $R$:

\begin{equation}
m_{v}^{t+1}= \mathrm{AGG}_{w \in N(v)} [M_{t} (h_v^t, h_w^t, e_{vw})]
\end{equation}

\begin{equation}
h_v^{t+1} = U_t(h_v^t, m_v^{t+1})
\end{equation}

\begin{equation}
\hat{y} = R(\{h_v^T | v \in G \})
\end{equation}

Of the MPNNs, we draw on the design of the Graph Convolutional Network (GCN) \cite{kipf2016semi} and ideas from \cite{xu2018powerful} to produce an architecture that performs well at the task of deciding whether a weighted graph should be divided into one or more partitions. The  message passing function for our proposed approach, the MAG-GCN (Mixture AGgregation Graph Convolutional Network), is

\begin{equation}
m_{v}^{t+1}= (\mathrm{MEAN}(h_{w}, e_{vw}), \mathrm{MAX}(e_{vw}), \mathrm{MIN}(e_{vw})),
\end{equation}
where $(\cdot, \cdot)$ denotes concatenation.  We found the concatenation of the edge feature to the node features be important, as observed in \cite{dwivedi2020benchmarking}.  Moreover, the inclusion of the min and max aggregation represents a significant departure from the message passing function of the GCN: $m_v^{t+1} \sim  \mathrm{MEAN}(h_{w}$).  We found that for reasoning about partitioning in a weighted graph, it was helpful for the message passing function to carry more information about the distribution of incoming edge weights than just the mean.  To demonstrate these points, we compare the MAG-GCN to an architecture that we refer to as GCN-E \cite{dwivedi2020benchmarking}, which is the same but with message passing 

\begin{equation}
m_v^{t+1} =  \mathrm{MEAN}(h_w, e_{vw}).  
\end{equation}
We also tried adding a sum aggregator to our mixed aggregation, but found that this did not yield any benefit.  We also note that for the graphs studied here, where all of the information is contained in the edge weights, it is not obvious how to adopt a vanilla GCN architecture such as that in \citet{kipf2016semi}, where the focus is on learning from node features.  After completion of this work, we became aware of \citet{corso2020principal}, which demonstrates that a similar mixed aggregation scheme enhances GNN performance in a broad array of applications.  



For all of the GNN architectures we studied, the update function consists of a fully connected neural network $f_U$ with ReLU activation, and the readout function consists of a fully connected network that operates on the mean of the hidden states $\bar h_{v} $, $R = f_R(\bar h_{v})$.  Following readout the loss is calculated with binary cross-entropy.  We utilized learning rates between $10^{-2}$ and $10^{-3}$.  Initial node features were set to a vector of zeros and the edge weights $e_{vw}$ were set to the pairwise edge similarity.  Hyperparameters were tuned on synthetic datasets.  Our graph neural network models were built with Deep Graph Library \cite{wang2019dgl}, a general framework for building Message Passing Neural Networks.


\begin{table*}[]
  \centering
  \caption{Performance on real and synthetic data.}
\begin{tabular}{llllll}
\hline
Dataset                    & TC             & K-Core         & SuperPart      & GCN-E         & MAG-GCN (Ours) \\
\hline
Softlines                  & 0.716 (0.006) & 0.624 (0.014) & \textbf{0.721} (0.005) &  0.705 (0.005)             & \textbf{0.721} (0.009) \\
Hardlines                  & 0.873 (0.008) & 0.663 (0.002) & 0.880 (0.004) &   0.877 (0.003)            & \textbf{0.887} (0.003) \\
IC-Stratified              & 0.725 (0.012) & 0.746 (0.026) & 0.754 (0.023) & 0.757   (0.003)           & \textbf{0.774} (0.003) \\
$Beta$(1,1), $Beta$(3,1.5) & 0.375 (0.047)  & 0.433 (0.041)  & 0.619 (0.071)  & 0.576 (0.061) & \textbf{0.687} (0.061)  \\
$Beta$(2.5,4), $Beta$(4,2) & 0.706 (0.057)  & 0.640 (0.067)  & 0.819 (0.04)   & 0.737 (0.039) & \textbf{0.843} (0.039)  \\
$Beta$(1,2), $Beta$(4,1.5) & 0.624 (0.051)  & 0.762 (0.051)  & 0.914 (0.028)  & 0.813 (0.048) & \textbf{0.932} (0.029)  \\
$Beta$(1.5,4), $Beta$(3,1) & 0.907 (0.03)   & 0.856 (0.04)   & 0.95 (0.016)   & 0.825 (0.037) & \textbf{0.967} (0.014)  \\
$Beta$(1.5,4), $Beta$(4,1) & 0.936 (0.035)  & 0.924 (0.028)  & \textbf{0.976} (0.019)  & 0.907 (0.036) & 0.973 (0.009)  \\
$Beta$(1,4), $Beta$(4,1)   & 0.979 (0.01)   & 0.972 (0.016)  & 0.981 (0.02)   & 0.93 (0.025)  &  \textbf{0.989} (0.014) \\
\hline
\multicolumn{5}{l}{We report mean $F_{1}$ score with parentheses enclosing standard error.}
\end{tabular}
\label{table:results}
\end{table*}

\begin{figure*}[t]
  \centering
  \includegraphics[width=6.25in]{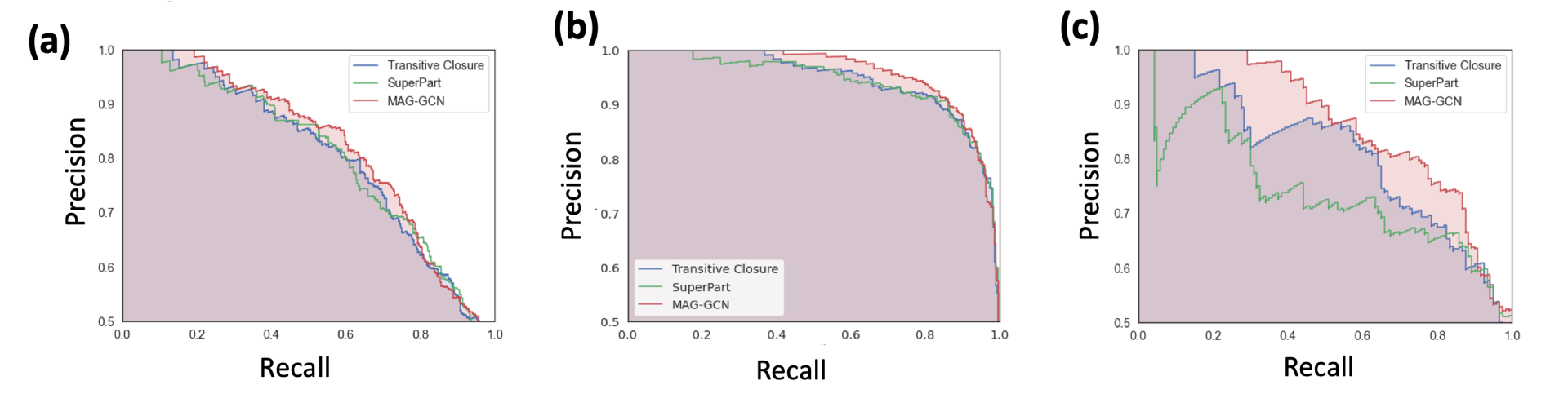}
  \caption{Performance of cluster scoring on: (a) Softlines , (b) Hardlines, (c) An open dataset introduced in \cite{reas2018superpart} and referred to in that work as IC-Stratified. F1 scores for the same datasets can be found in Table \ref{table:results}.}
    \label{fig:results}
\end{figure*}

%

\section{Results}

\begin{figure*}[t]
  \centering
  \includegraphics[width=6.25in]{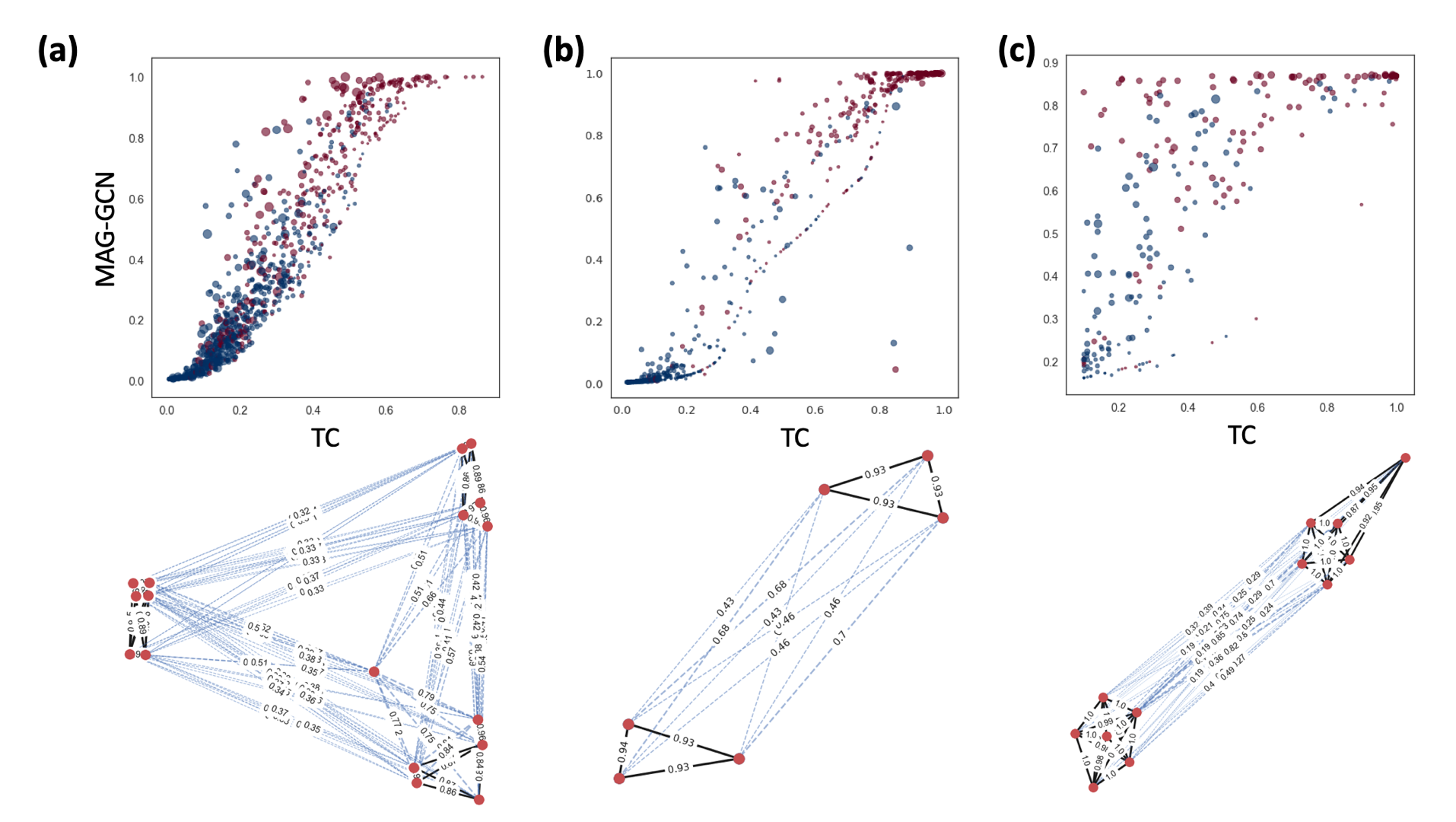}
  \caption{Study of improved performance.  Upper: MAG-GCN score is displayed against the TC score for (a) Softlines, (b) Hardlines, (c) IC-Stratified.  Blue points are ground truth negative ICs, red points are ground truth positive ICs, and size represents the number of items in the family.  The upper left hand corner of the graph represents the area where the GCN increased the score compared to TC.  The primary effect of the GCN is to boost the scores of positive examples more than negative examples.  Lower:  example families of products that were classified as non-IC by TC but were correctly classified as IC by MAG-GCN.  The red points represent items, black edges represent true positive edges, and blue dotted lines represent true negative edges.  The width of the line is proportional to the pairwise model prediction $p$.  For example, in the rightmost plot, owing to a single edge with weight 0.85 between two true clusters, TC did not suggest partitioning into two families with high likelihood (score = 0.15); however, MAG-GCN correctly did (score was 0.78, true label was 1.0)}
  \label{fig:analysis}
\end{figure*}

\begin{figure*}[t]
  \centering
  \includegraphics[width=6.25in]{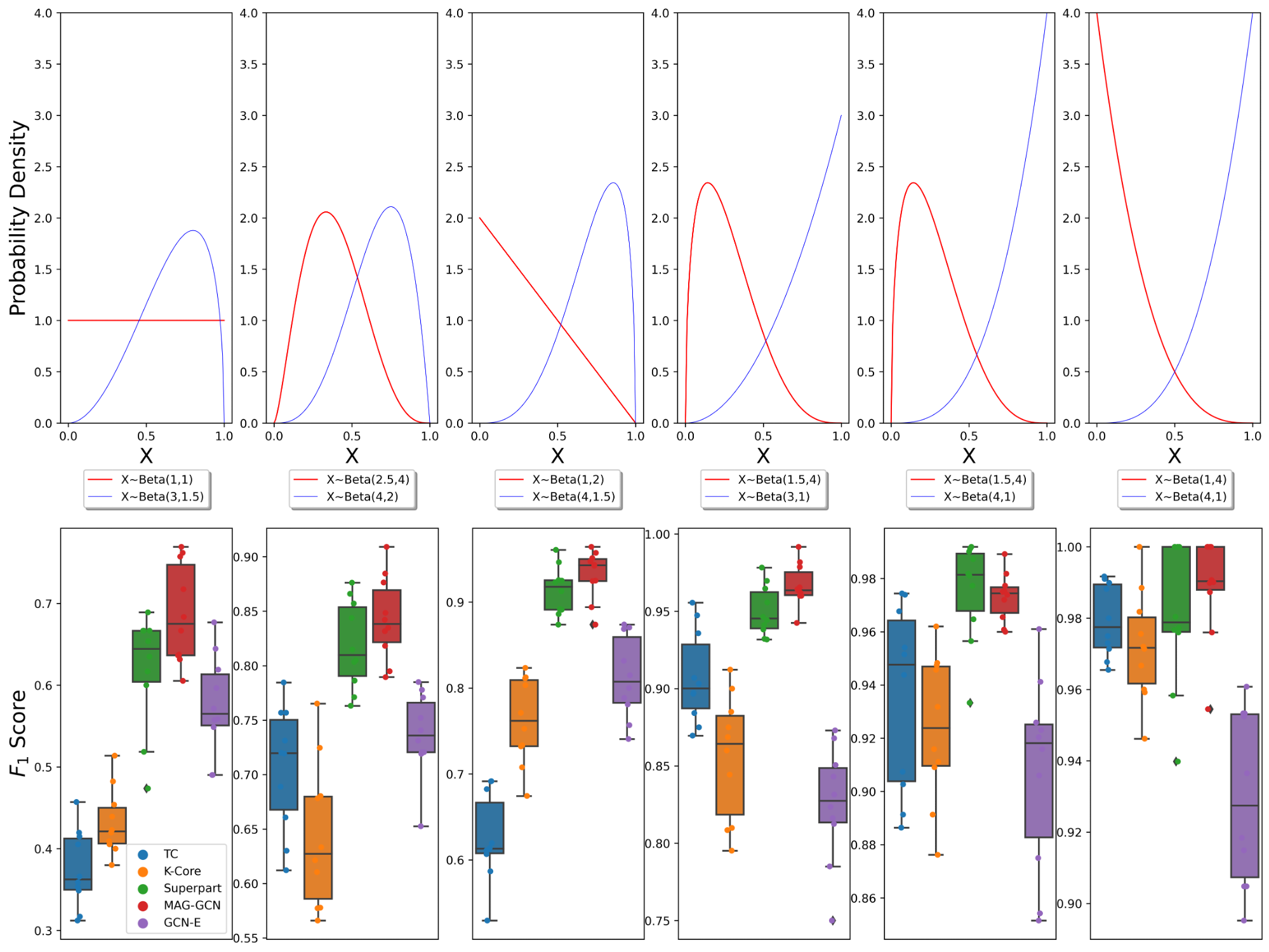}
  \caption{Performance on synthetic data.  Top: beta distributions used generate synthetic data.  Negative edges are drawn from the red distributions while positive edges are drawn from the blue distribution.  Bottom:  different cluster scoring techniques are applied to each distribution.  We utilize the synthetic datasets from Table \ref{table:results} (N = 1000).  }  
  \label{fig:synthetic_combined}
\end{figure*}

\subsection{Experimental design}

To evaluate model performance, we choose transitive closure, k-core, and SuperPart as baseline methods. For evaluating performance on a test set, we use F1 score.  For the F1 score metric, a threshold is first chosen to maximize F1 score on a validation set, and the F1 score is evaluated on a test set.

We first evaluated all of the methods in an internal application. Several audited datasets are created as follows. A group of items was shown to auditors on a single page, and the auditors were given the task of partitioning it into one or more Variation families. One audited dataset consists of approximately 8,500 such families from a Softlines product line containing 76,000 items which have been partitioned according to auditor labels; approximately 3,800 of these families were split by the auditors and are therefore IC.  The second audited dataset consists of approximately 6,500 families from a Hardlines product line containing 124,000 items; 2,700 families are IC.  

We also use a publicly available dataset released in \citet{reas2018superpart} called IC-Stratified; this dataset contains a training set with 2649 nodes and a test set with 2424 nodes with ground truth partitionings into one or more families. 

Finally, we evaluated our methods on synthetic datasets. When generating the datasets, we tested different degrees of separation between positive and negative pairs, as well as the specific distributions for generating the weights of these pairs. The systematic study on synthetic datasets allow us to gain deep insights on the advantages and disadvantages of the different methods under different conditions.  We generate synthetic data  by creating graphs of 1 or 2 ground truth families, where the latter is an IC example. Each node had probability of 0.97 to be part of the first true family and 0.03 to from the second. All edges within true families were drawn from one Beta distribution, and all edges between true families were drawn from another.

\subsection{Real-World Datasets}

For the Softlines and Hardlines datasets, we trained and tested the models on a fixed training, validation, and test set 5 times, each with a different random seed for the pairwise classifier training.  For IC-Stratified, pairwise scores are provided and we split the training set into training and validation 5 different ways.  A comparison of the cluster scoring models can be found in Table \ref{table:results}, with PR-AUC curves shown in Figure \ref{fig:results}.  Overall, we find that a model-based cluster scoring mechanism helps with the task of identifying inconsistent clusters, with both SuperPart and MAG-GCN outperforming the TC baseline.  We find that the MAG-GCN model performs the best overall in Softlines, Hardlines, and the public dataset IC-Stratified from \cite{reas2018superpart}.  The GNN model is particularly effective at improving recall at high precisions, an important metric for applications.  An example precision-recall comparison between the MAG-GCN and the TC baseline can be found in Figure \ref{fig:results}.

To understand the origin of the improved performance, we plot the GNN score against the TC score in Figure \ref{fig:analysis}.  We observe that the MAG-GCN derives its improved performance by boosting the scores of TC's false negatives more than reducing the scores of false positives.  We also observe that larger families tend to benefit more often from the GNN than their smaller counterparts.  This pattern can be better understood by examining some examples.  The lower portion of Figure \ref{fig:analysis} shows example families that were correctly labeled positive by the GCN but incorrectly by transitive closure.  We see that while the pairwise model correctly drew low-probability edges between clusters in many cases, there is often a structural component in the graph that makes it difficult to partition by transitive closure.  For example, in the example from Softlines (a), there is a central node that holds three separate clusters together.  In the examples from Hardlines (b) and IC-Stratified (c) there are widely varying edges between two true clusters.

\subsection{Synthetic datasets}
A study of synthetic data is shown in Figure \ref{fig:synthetic_combined}.  As mentioned previously, we produced artificial families of items ranging from size 5 to 15 using a different distribution for true positive edges and true negative edges.  By varying the parameters of the distribution, we were able to determine how methods of cluster scoring perform across different  levels of difficulty.  The distributions and corresponding performances are shown in Figure \ref{fig:synthetic_combined} and Table \ref{table:results}.  We find that MAG-GCN consistently outperforms or is competitive with the other methods, with particular performance increases for noisy data.  For example, when positive edges are drawn from $Beta(3,1.5)$ and negative edges are drawn from a uniform distribution -- a noisy setting -- MAG-GCN has an F1-score of 0.687 vs. 0.375 for TC and 0.619 for Superpart - an $83 \%$ and $11 \%$ improvement over the baselines, respectively.  However, when both positive and negative distributions are well-separated, most methods do well.  This points to the usefulness of our method in the noisy data setting, which is common in applied settings such as our internal task.

We also tested all methods against the GCN-E, which is similar to MAG-GCN but uses only mean node aggregation rather than the mixture of mean, min, and max.  We note that the concatenation of the edge feature into a GCN alone is insufficient to learn this task as successfully as our method.


\section{Discussion}
As in previous work \cite{reas2018superpart}, we find that by applying an additional cluster scoring model we can improve performance on the joint task of classifying an entire family as inconsistent.  Of the cluster scoring mechanisms we tested, we find that the MAG-GCN is most effective, outperforming its immediate competitor SuperPart.  We also found that the benefit of the MAG-GCN is most dramatic for recall in the high precision regime, which is a relevant metric for production use cases where we wish to fix only high-confidence IC families.  The performance of the MAG-GCN in comparison to SuperPart was also interesting.  We noticed that when little data was available, SuperPart performed well and once better than MAG-GCN (second to last plot in Figure \ref{fig:synthetic_combined}), suggesting that the SuperPart model may be better regularized.  However, MAG-GCN appears to be more flexible and able to learn effective representations of a wide variety of weighted graphs.  Understanding the power of Graph Neural Networks relative to featurized representations of graphs is an interesting topic for further study.

We also make note that MAG-GCN is significantly better than GCN-E, where the edge feature is simply concatenated and the aggregation function is the mean.  Xu et al. \cite{xu2018powerful} note the mean aggregation function is effective at learning the distribution of its surroundings while the max aggregation function is more effective at distinguishing outliers.  We note that the mean aggregation function provides information about the center of the distribution of incoming edges, while the min and max aggregation functions provide some sense of its spread.  

To frame the results in terms of our IC detection application, the decision at the cluster level, rather than pairwise, appears to add value over TC primarily by increasing model score for families that would be false negatives under TC because they are tied together by outlier edges.  In Figure \ref{fig:analysis}, the benefit of the GNN score appears to increase for larger families, where there is more opportunity to have an outlier edge.  We suspect that in real-world applications of inconsistent family detection to product graphs, where there may be hundreds or even thousands of products linked together, the benefits of this technique will grow.  

Despite the performance enhancements achieved by the MAG-GCN, further work including raw node features such as images and text is needed to take full advantage of the graph neural network paradigm for product relationships.  For the use case of automatically cleaving high confidence families, further work is recommended on using GNNs to learn the correct partitioning within the IC.  Models whose loss depends on the categorization of each node such as the Line Graph Neural Network \cite{chen2018supervised} might be well suited for this task.  

\section{Conclusions}
Using graph neural networks that leverage graph structure, we are able to significantly improve the ability of cluster scoring models to detect Inconsistent Clusters for e-commerce applications.  More generally, we show that graph neural networks can learn robust graph representations of weighted graphs for partitioning decisions.  Future work will focus on incorporating additional node and edge features, as well as studying graph neural network structures that partition the graph explicitly rather than classify the entire graph as either consistent or inconsistent.   For the binary classification problem, we hypothesize that methods allowing learnable coarsening of the graph \cite{swietojanski2015differentiable, cangea2018towards} could be useful.  Graph neural networks offer an interesting path forward for other entity resolution problems because node features can be included along with the edge weights studied here.

In terms of incremental entity resolution, focusing on the IC problem allows us to handle the important case where we want to enforce that cluster quality does not degrade with ER updates.  Having an Inconsistent Cluster detection mechanism is an efficient way to accomplish this regardless of the quality of pairwise scores, since precision and recall for cluster-level updates can be tuned along the curves shown in Figure \ref{fig:results}.  In this paper, we mainly studied the IC detection problem in the context of existing clusters prior to splitting, but the same cluster quality checks can be applied to clusters prior to being merged or split.  Future work will include evaluating how to combine merging and splitting to optimize the IER in the general case.  

\bibliographystyle{ACM-Reference-Format}
\bibliography{gnn_bibliography}

\end{document}